# Automated stance detection in complex topics and small languages: the challenging case of immigration in polarizing news media

Mark Mets[1,3]*, Andres Karjus[1,3]*, Indrek Ibrus[2,3], Maximilian Schich[2,3]

[1] School of Humanities, Tallinn University, Estonia
[2] Baltic Film, Media and Arts School, Tallinn University, Estonia
[3] ERA Chair for Cultural Data Analytics, Tallinn University, Estonia
* Corresponding authors: mark.mets@tlu.ee, andres.karjus@tlu.ee

Automated stance detection and related machine learning methods can provide useful insights for media monitoring and academic research. Many of these approaches require annotated training datasets, which limits their applicability for languages where these may not be readily available. This paper explores the applicability of large language models for automated stance detection in a challenging scenario, involving a morphologically complex, lower-resource language, and a socio-culturally complex topic, immigration. If the approach works in this case, it can be expected to perform as well or better in less demanding scenarios. We annotate a large set of pro and anti-immigration examples, and compare the performance of multiple language models as supervised learners. We also probe the usability of ChatGPT as an instructable zero-shot classifier for the same task. Supervised achieves acceptable performance, and ChatGPT yields similar accuracy. This is promising as a potentially simpler and cheaper alternative for text classification tasks, including in lower-resource languages. We further use the best-performing model to investigate diachronic trends over seven years in two corpora of Estonian mainstream and right-wing populist news sources, demonstrating the applicability of the approach for news analytics and media monitoring settings, and discuss correspondences between stance changes and real-world events.

## Introduction

Understanding complex socio-political and cultural issues, such as polarization and news biases requires a comprehensive perception of cultural systems. Computational and data-driven research can offer valuable insights, provided we acknowledge the limitations of computational methods and scrutinize findings. Advances in natural language processing, such as pretrained large language models (LLMs) have enabled the analysis of large volumes of data, but these methods may have limited applicability in smaller languages with limited training data and NLP resources. This includes dealing with politically charged issues that involve diverse linguistic expressions and cultural perspectives. However, quantifying the reporting of different arguments or stances towards various issues can help scholars to better understand media ecosystems, study the political positions of different media groups or specific outlets, but can also aid industry, including media organizations if they are looking to balance and avoid bias in their reporting.

We report on an experiment of automatically classifying topic-specific political stance in news media texts written in a low to medium-resource, morphologically complex language, Estonian, spoken natively by about 1.1 million people, primarily in the European state of Estonia. While we use one small language as the example, we argue below that our results have implications for the applicability of automated stance detection and media monitoring more broadly. The topic in question is the globally much disputed and often polarizing topic of immigration. Our corpus consists of news articles published in 2015-2022 by one mainstream media group (Ekspress Grupp), and one right-wing populist online news and opinion portal (Uued Uudised, or the "new news"). The study is based on an academia-industry collaboration project with the Ekspress media group, who provided data from their in-house publishing database (but did not influence the design of the study nor the conclusions). Their interest was to assess the neutrality of their content. The goal of this study was to determine the feasibility and accuracy of automated stance detection for linguistically and culturally complex issues (in this case, immigration), in a lower-resource language, and also apply it to mapping stance in a large corpus of news dealing with the topic. This could be applied to assess the balance of different views in news reporting as well as to foster discussions about bothsidesism. For that purpose we compare sources that may likely have contrasting views on the politically charged topic of immigration. We focus on testing a supervised learning approach, annotating a set of training data, tuning a number of different LLMs on the training examples, and testing them on a holdout test set. The best-performing model is further applied to the larger corpus to estimate the balance of different stances towards immigration in the news.

Our experiment design follows a fairly standard annotate-train-test procedure. We first extracted 8000 sentences from the joint corpus using a lexicon of topic-relevant keywords and word stems (referring to keywords such as *migrant*, *immigration*, *asylum seeker*), carried out manual stance annotation, and fine-tuned a number of pre-trained LLMs on this dataset for text classification, including multilingual and Estonian-specific ones. We also experiment with zero-shot classification in the form of instructing ChatGPT 3.5 to classify sentences according to similar guidelines as the human annotators. All LLM classifiers achieve reasonably good test set accuracy, including the zero-shot variant, which performs almost on par with the best annotations-tuned model. Our work has three main contributions:

We demonstrate the feasibility and example accuracy of what amounts to a proof of concept for an automated political stance media monitoring engine, and also compare it to cheaper approaches of bootstrapping a general sentiment analysis classifier to estimate stance, and using zero-shot learning. While not perfect, we argue the approach can yield useful results if approached critically, keeping the error rates in mind. We have chosen a socio-politically complex example topic and a lower-resource language for this exercise. Consequently, it is reasonable to expect higher accuracy when following analogous procedures, where one or more of the variables are more favorable: either the target language having larger pre-trained models available, the topic being of lesser complexity, or larger quantities of training data are annotated. We are making our annotated dataset of 7345 sentences public, which we foresee could be of interest to the Estonian NLP as well as media and communications studies communities, as well as applications of multilingual NLP and cross lingual transfer learning. We also contrast the more traditional annotations based training approach to zero-shot classification based on using an instructable LLM, ChatGPT. We offer a perspective of such an approach's future importance in academia and beyond. While first attempts at testing ChatGPT as a zero-shot classifier have focused on large languages like English, we provide insight into its performance on a lower-resource language.

Secondly, we carry out qualitative analysis of the annotation procedure and model results, highlighting and discussing difficulties for both the human annotators and the classifier, when it comes to complex political opinion, dog-whistles, sarcasm and other types of expression requiring contextual and cultural background knowledge to interpret. Lessons learned here can be used to improve future annotation procedures.

Finally, we show how the approach could be used in practice by media and communications scholars or analytics teams at news organizations, by applying the trained model to the rest of the corpus to estimate stances towards immigration and their balance in the two news sources over a 7 year period. This contributes to understanding immigration discourse, media polarization and radical-right leaning media on the example of Estonia. The topic is also an example of real-world commercial interest, where our industry partner has been interested in keeping balance of their reporting of different stances. We find and discuss qualitative correspondences between changes in stance and relate them to events such as Estonian parliamentary elections in 2019 and the start of the Russian invasion to Ukraine of 2022.

**Analytic approach**

We approach stance detection as determining favorability toward a given (pre-chosen) target of interest (Mohammad et al., 2017) through computational means. Stance detection (or stance classification, identification or prediction) is a large field of study, partially overlapping with opinion mining, sentiment analysis, aspect-based sentiment analysis, debate-side classification and debate stance classification, emotion recognition, perspective identification, sarcasm/irony detection, controversy detection, argument mining, and biased language detection (ALDayel & Magdy, 2021; Küçük & Can, 2020). Stance detection is used in natural language processing, social sciences and beyond in order to understand subjectivity and affectivity in the forms of opinions, evaluations, emotions and speculations (Pang & Lee, 2008). Compared to sentiment analysis, which generally distinguishes between positivity or negativity, stance detection is a more topic-dependent task that requires a specific target (Mohammad et al., 2017) or a set of targets (Sobhani et al., 2017; Vamvas & Sennrich, 2020). The distinction is of course not clearly categorical, with e.g. aspect-based sentiment analysis, commonly applied to product reviews, being applicable to multiple targets (Do et al., 2019). We chose to assess stance towards one target, immigration, and contrast our results with using an existing Estonian sentiment analysis dataset to fine-tune a classifier based on the best performing LLM.

Both sentiment analysis and stance detection are classification tasks with multiple possible implementations. Earlier approaches were based on dictionaries of e.g. positive and negative words, and texts would be classified by simply counting the words, using rules of categorization, or various statistical models. We employ the method of tuning large pretrained language models like BERT (Devlin et al., 2019) as supervised text classifiers. Such context-sensitive language models have been shown to work well across various NLP tasks and typically outperform earlier methods (Devlin et al., 2019; Ghosh et al., 2019). Reports on using LLMs for stance detection in lower-resource languages are relatively limited in literature. However, their usefulness is clear in scenarios where language-specific NLP tools and resources such as labeled training sets may be lacking, but there is enough unlabeled data such as free-running text to train a LLM or include the language in a multilingual model (Hedderich et al., 2021; Magueresse et al., 2020). Resources relevant for NLP include both available methods as well as datasets among other factors (cf. Batanović et al., 2020).

Automated stance detection has also been relevant in studies on immigration and related topics. The data they use is most often textual, ranging from often studied Twitter (ALDayel & Magdy, 2021; Khatua & Nejdl, 2022) to online discussion forums (Yantseva & Kucher, 2021) and comments of online news (Allaway & McKeown, 2020). In the context of news media, the immigration topic is also relevant in hate-speech detection, which applies similar methods (Khatua & Nejdl, 2022). These studies use a variety of methods for stance detection, including LLMs. These include single-shot studies (e.g. Card et al., 2022) where training set topics match the predicted topics; multi-shot approaches which offer partial transferability; and zero shot (Allaway &

McKeown, 2020; Vamvas & Sennrich, 2020) which aims to predict topics not contained in the training set. Automated stance detection has been used to study immigration topics in under-resourced languages (Yantseva & Kucher, 2021), and across-topics (zero-shot) and multilingual approaches using LLMs have been shown to work across languages other than English like Italian, French and German (Vamvas & Sennrich, 2020).

**Object of analysis**

Immigration has witnessed increased focus in media and politics in Europe since the 2015 European migrant crisis, but is also relevant globally. Analysis of media representations of immigration is crucial, as it can determine stances towards immigration (Burscher et al., 2015; Meltzer & Schemer, 2021), such as perception of the actual magnitude of immigration. In turn, exposure to immigration related news can have an impact on voting patterns (Burscher et al., 2015). This topic is also central in populist radical right rhetorics (Mudde, 2007; Rooduijn et al., 2014). Social media has been argued to be one of the means for achieving populistic goals (Engesser et al., 2017: 1122). In the Estonian context, most of the radical right content circulating in Estonian-language social media have been reported to be references to articles from the news and opinion portal Uued Uudised (Kasekamp et al., 2019), making it a relevant source for understanding radical right populists' perspective towards immigration.

Focus on immigration fits into populistic rhetoric in the context of distancing the "us" from the strange or the "other". In the case of the radical right, this other may be often chosen based on race or ethnicity (Abts & Rummens, 2007: 419). Such exclusionism of immigrants and ethnic minorities can be present in radical right populism to the extent that it becomes its central feature (Mudde, 2007; Rooduijn et al., 2014). Who that minority group is varies and may also change over time. For example, before 2015, Central and Eastern European (CEE) populist radical right parties used to target mainly national minorities, whilst in Western-Europe it was more often immigrants. After the 2015 immigration crisis, immigrants also became the main target in the CEE countries (Kasekamp et al., 2019).

The same applies to Estonia, where immigration has been one of the topics used by the radical right parties to grow their political impact, especially since 2015 (Kasekamp et al., 2019). 2015 also marked the emergence of many anti-immigrant social media groups, blogs and online news and opinion portals which have gained popularity since then. This includes the radical right online news portal Uued Uudised, a channel whose news are often ideologically in line with and give voice to the political party of EKRE ("Conservative People's Party of Estonia"). In academic literature EKRE has often been classified as a radical right populist party (Auers, 2018; Kasekamp et al., 2019; Koppel

& Jakobson, 2023; Madisson & Ventsel, 2018; Petsinis, 2019) whilst the party describes itself as national conservative (Madisson & Ventsel, 2018; Saarts et al., 2021). Uued Uudised has been described as both alternative (Kasekamp et al., 2019) as well as hyper partisan media (Saarts et al., 2021). It was established in 2015 during the EU immigration crisis. The content of the Estonian radical right media discourse is often following provocative and controversial argumentations (Kasekamp et al., 2019). Immigrants are often constructed as an antithetical enemy, where the Other is portrayed as a mirror image of the Self, whereas the Other may first be given negative characteristics that are then perceived as nonexistent in one's own group (Kasekamp et al., 2019; cf. Lotman et al., 1978; Madisson & Ventsel, 2016). Such othering towards immigration can also be noticed through the topics discussed in the media more broadly, such as framing immigration in the context of criminal activity (Kaal & Renser, 2019; Koppel & Jakobson, 2023).

While our study has a methodological focus, we have chosen an example that also contributes to a better understanding of the topic of immigration, media polarization and radical-right discourse in our example country of Estonia. Radical-right discourse has been an under-researched topic in Estonian context (Kasekamp et al., 2019). Political science has focused on communication of the parties themselves (Braghiroli & Makarychev, 2023; Petsinis, 2019; Saarts et al., 2021), while textual analyses have often focused on social media (Kasekamp et al., 2019; Madisson & Ventsel, 2016, 2018). These qualitative studies can benefit from a large-scale quantitative approach through automated stance and sentiment detection offering a complementary perspective.

## Methods and materials

### Dataset

We chose the data based on accessibility, and to contrast two sources expected to have different stances on immigration. The corpus consists of articles from 2015 to the beginning of April 2022. The mainstream news are from the Ekspress Grupp, one of the largest media groups in the Baltics. Our data covers one dominant online news platform, Delfi, across all of the time period, and a sample from multiple other daily and weekly newspapers and smaller magazines. The populist radical-right leaning media is represented by the abovementioned online news portal Uued Uudised.

We acquired the Ekspress Grupp data directly from the group and scraped Uued Uudised from its web portal. Both datasets were cleaned of tags and non-text elements. We included Estonian language content only (the official language of the country is Estonian, but there is a sizable Russian speaking minority, and both news sources include Russian language sections). Our dataset consists of 21 667 articles from Uued Uudised (April 2015 to April 2022) and 244 961 from Ekspress Group (January 2015 to March 2022). The received data of Ekspress Group was incomplete with a gap in October-December 2019. The data from 2020 onwards contains multiple times more content from other periodicals besides Delfi (cf. Fig S1 for detailed distribution of Ekspress data).

We chose sentence as the unit of analysis, instead of e.g. paragraph or article, for three reasons. The length of articles varies greatly, as does the length of a paragraph across articles, and some articles lack paragraph splits. Secondly, longer text sequences may include multiple stances, which may confuse both human annotators and machine classifiers. Thirdly, the computational model, BERT, has an optimal input length limit below the length of many longer paragraphs. It was hoped a sentence would be a small enough unit to represent a single stance on average, but enough context to inform the model. Admittedly, sentence level analysis does have the limitation of missing potentially important contextual information across sentences, as we further discuss in our annotation and classification-error analysis. It is often hard to deduce an opinion from a single sentence length text alone (cf. Mohammad et al., 2017), but we do expect sentence to be a suitable unit of analysis to indicate changes in rhetoric and large-scale changes across time.

We extracted immigration-related sentences using a dictionary of keywords to cover different aspects related to immigration, implemented as regular expressions (also to account for the morphological complexity of Estonian and match all possible case forms). Previous research on immigration has approached sampling by choosing topic-specific datasets, like immigration related discussion forums (Yantseva & Kucher, 2021) or using dictionary based approaches, like Card et al. (2022). We found using predefined keywords as simple and efficient enough for our task. Using text embeddings can provide a good alternative if keywords are harder to limit or have many synonyms (Du et al., 2017). We created a list of keywords sorted into groups representing various aspects of the migration as well as other closely related topics — migration, refugees, foreign workers, foreign students, non-citizens, race, nationality, and terms related to radical-right and liberal opposition (e.g. "multiculturalism") terms (cf. Appendix for more information on keywords and Fig. S2 for distribution of keywords groups). This plurality of topics (e.g. also covering "digital nomads") made the task much more challenging but at the same time allowed to grasp more nuances of the migration discourse at large. This yielded sentences that included both opinions as well as factual descriptions and were therefore stylistically varied. In addition to searching for relevant keywords, we used a negative filter to exclude unrelated topics, like bird migration (see Fig 1 for distribution of filtered sentences).

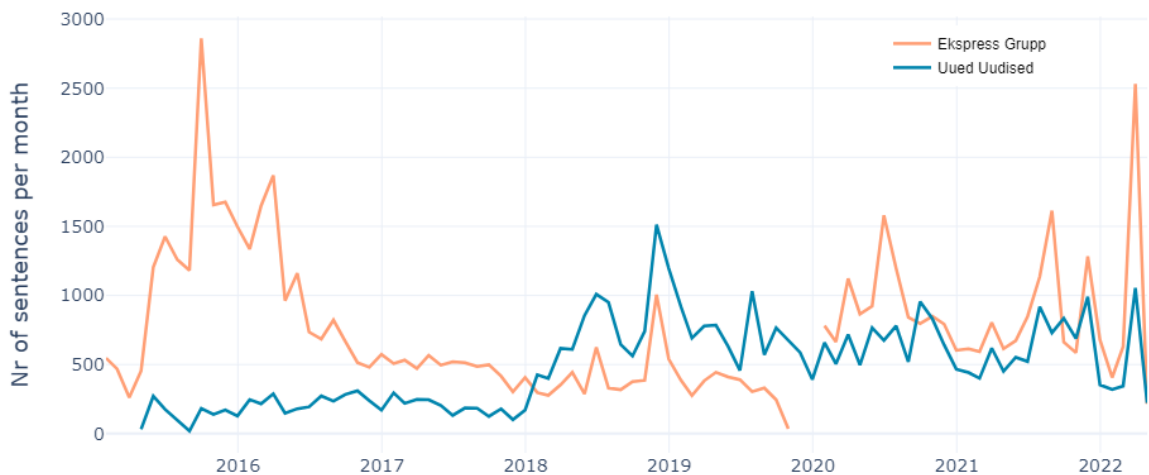

**Figure 1. Monthly distribution of immigration related sentences**. The red line represents the Ekspress Grupp and blue Uued Uudised. There is no data for Ekspress Grupp at the end of 2019, where the count is 0. The change of relevant sentences in Ekspress Grupp after that reflects the difference in the dataset, which was larger and was more varied in terms of specific periodicals (cf. Figs, S1 and S3 for distribution of articles per Ekspress Grupp periodical and Fig. S4 for similar distribution of immigration related sentences but per week).

**Annotations**

We assigned two Estonian speaking graduate students to annotate a total of 8000 sentences for supervised training. The annotators were compensated monetarily. The sample was balanced by keyword prevalence and publishers (Uued Uudised and Ekspress Grupp), but not by the time or source article. Based on annotator feedback, we removed very long, repetitive or list-like and non-topical sentences, leaving 7345 sentences. The sentences were annotated on a 1-5 point scale from pro- to anti-immigration, with the option to mark the sentence as ambiguous instead. Ratings were later reduced to four classes of Against (1-2), Neutral (3), Supportive (4-5), and Ambiguous. The latter was meant for sentences that we assumed to be unhelpful for model tuning, and were therefore excluded, including sentences that were either unintelligible, non-topical or expressed multiple stances at once. While some sentences are straightforward to interpret, others can pose a challenge for annotation due to complex metaphorical usage or references requiring additional knowledge. Below are some examples, translated into English (original Estonian in the Appendix).

1) "Mass immigration would be disastrous for Europe and it would not solve anything in the world." (Against)
2) "The process to get a residence permit here was not very complicated." (Supportive)
3) "Migration issues must definitely be analyzed, including the aspect of international obligations and their binding nature, and various steps should be considered." (Neutral)
4) "One can only wonder - when do Libyans quit and follow the flow of things when Europe is just talking about controlling the migrant crisis but itself just pours oil on fire." (Ambiguous)
5) "It is not worth mentioning that the person in question is thoroughly Europhile and globalist." (Against, because the manner presumes that it is said from the perspective of someone who may be against immigration)
6) "They criticize racism, homophobia, xenophobia and what they see as outdated nationalism." (Supportive, but refers to a third person and may thus also be taken as Against)

The sentiment analysis classification used for comparison differed from stance detection only in terms of the annotations used for fine-tuning. We used a publicly available Estonian language dataset of short paragraphs labeled for sentiment as Negative, Neutral, Positive and excluded Mixed class (Pajupuu et al., 2016).

A third annotator (the first author) later annotated a subset from both of the previous annotators to estimate inter-rater agreement. There was substantial agreement on Supportive, Again and Neutral classes (κ = 0.69 and 0.66 between the third and each of the other annotators) (see Appendix for details). There was a very strong agreement between only Pro and Against classes (κ = 0.97 for both), indicating that most of the disagreement was between one extreme and Neutral.

## Automatic classifiers

We use our annotated dataset to train and compare several popular LLMs based on BERT and BERT-like (Devlin et al., 2019) transformers architecture: multilingual mBERT (Devlin et al., 2019), XLM-RoBERTa (Conneau et al., 2020), monolingual EstBERT (Tanvir et al., 2021) and Est-RoBERTa (Ulčar et al., 2021). We used the larger versions of the publicly available models with 512 tokens to fit longer sentences that were optimal for our setup. We used a Simple Transformers library in Python (https://github.com/ThilinaRajapakse/simpletransformers) for working with transformers. We started by estimating best hyperparameters for each model, minimizing evaluation loss (see Appendix for hyperparameters), and then measured model performance with cross validation. The evaluation set was 20% of data in each case. As the training and test data was unbalanced in terms of the number of classes, our training took into account the weights (relative size) of the classes.

In addition, we also compared the results with that of GPT-3.5 based ChatGPT (we conducted our experiments on March 3, 2023, using the February 13 version of ChatGPT 3.5). It is a more recent large language model specifically trained for dialogue. The new approach of using (even larger) generative LLMs as zero-shot classifiers (also known as prompt-based learning, cf. Liu et al., 2023) has opened up potentially new avenues of cheap and efficient text classification, as it requires no fine-tuning and can simply be instructed using natural language.

There has been a surge of research on ChatGPT performance for different NLP tasks, but mostly focused on English. It has been shown that ChatGPT can achieve similar or better results in English than comparable supervised and other zero-shot models, including in stance detection (Zhang et al., 2023). On a wider array of tasks, ChatGPT has been shown to be a good generalist model, but performing worse than models fine-tuned for a specific task (Qin et al., 2023). The model is problematic in terms of evaluation and replicability due to the ongoing development and closed nature of the model (Aiyappa et al., 2023). Our goal is to estimate its potential relevance for future studies by comparing it with the established pipeline of supervised tuning of pretrained LLMs for classification tasks.

We created a prompt that included optimized classification instructions and input sentences, in batches of 10. Responses not falling into Against, Neutral or Supportive classes were requested again until only labels belonging to this set were returned. Also if a wrong number of tags was returned, the sentences were requested again. An example input and output would look as follows:

**Input**: *Stance detection. Tag the following numbered sentences as being either "supportive", "against" or "neutral" towards the topic of immigration. "Supportive" means: "supports immigration, friendly to foreigners, wants to help refugees and asylum seekers". "Against" means: "against immigration, dislikes foreigners, dislikes refugees and asylum seekers, dislikes people who help immigrants". "Neutral" means: "neutral stance, neutral facts about immigration, neutral reporting about foreigners, refugees, asylum seekers". Don't explain, output only sentence number and stance tag.*
*1. Unfortunately, by now the violence has seeped from immigrant communities to all of the society.*
*2. [truncated]*
**Output**:
*1. Against*
*2. [truncated]*

# Results

The best-performing fine-tuned model was based on Est-RoBERTa, achieving an acceptable F1 macro score of 0.66 (precision 0.65 and recall 0.68; see Table 1). The difference with other monolingual EstBert (0.64) and multilingual XLM-RoBERTa (0.64) was minimal. All of the fine-tuned models performed better at classifying Against than Supportive stances. Est-RoBERTa model achieved F1 0.74 for Against, 0.69 for Neutral and 0.55 for Supportive class. The misclassification was mostly between Neutral and one extreme (see Fig. 2), similarly to e.g. Card et al. (2022). We regard it preferable to confusing the two extremes. The results are comparable to similar studies, and there is little difference between the models. Classifier trained on an existing sentiment dataset with Est-RoBERTa achieved the worst score, but performed better than expected. There were more mistakes between the two extremes than with sentiment analysis training set (see Annex for sentiment confusion matrix). We confirmed sentiment analysis training set performance by comparing the sentiment and stance predictions for all of the immigration related sentences, resulting in a fair agreement (kappa 0.29). It demonstrates the complexity of our task, which included features from stance as well as sentiment. Finally, comparable performance of zero-shot ChatGPT with the best model shows it could serve as a viable but cheaper alternative to fine-tuned models.

| Model | Against | Neutral | Supportive | F1 macro |
|---|---|---|---|---|
| Naive Bayes | 0.59 | 0.58 | 0.38 | 0.52 |
| EstBert class | 0.69 | 0.70 | 0.53 | 0.64 |
| **Est-RoBERTa** | **0.74** | **0.69** | **0.55** | **0.66** |
| XLM-RoBERTa | 0.73 | 0.65 | 0.54 | 0.64 |
| mBERT (cased) | 0.66 | 0.64 | 0.40 | 0.56 |
| mBERT (uncased) | 0.64 | 0.58 | 0.38 | 0.54 |
| ChatGPT (GPT 3.5) | 0.74 | 0.64 | 0.57 | 0.65 |
| Est-RoBERTa Sentiment | 0.63 | 0.42 | 0.42 | 0.49 |

**Table 1. Comparison of classification models.** F1 scores from different models by each class and across all classes. Bold indicates the best result with Est-RoBERTa. We used 5-fold cross-validation with 20% of data with all models.

**Figure 2. Confusion matrix of stance detection.** Based on one fold from the best performing model. Percentage shows the overlap between true (annotated) and predicted classes. Ideal but non-realistic classification would be 100% for diagonal from bottom left to top right. We regard the small values in top left and bottom right as a good sign, showing that most of the mistakes were between Supportive or Against, and Neutral, not between the two extremes (cf. Fig. S5 for comparison with sentiment analysis).

We further assessed the mistakes made by the best performing classifier. We looked at the mistaken predictions in the evaluation set between Against and Support classes and observed at least four types of interpretable mistakes.

1) Mistaken human annotations. These may be hard to fully exclude when using human annotations but could be reduced with better instructions.
2) Sarcasm, a well known challenge in NLP
3) Ambiguous and context dependent sentences. These may be generally more complicated to classify
4) Sentences that refer to a third person. These are tricky, as referencing someone else's opinion may implicitly imply agreeing or opposite standpoint, which is highly context dependent and therefore not easy, but a simpler task for humans than classifiers. These could relate to our chosen unit of analysis; paragraphs might perform better.

**Limitations**

The limitations of classifier performance are at least partly rooted in human annotations. Some of these shortcomings were reported by the annotators themselves. The distinction between neutral and ambiguous classification was also problematic, where more clear instructions might have helped. Confusion between neutral and ambiguous classes is not expected to have a strong effect on the results, but may have limited the size of our training set by having neutral sentences classified as ambiguous. Annotations are also dependent on the annotators' prior knowledge and biases. Annotators were instructed to only rate the sentence itself, but we expect that they also relied on personal contextual knowledge (cf. Batanović et al., 2020 for more aspects for annotators to consider in future studies). Yet, LLM are not impervious to (e.g. training set induced) biases either. It may explain why in some cases smaller and more specific models might perform better (Bender et al., 2021).

We suspect that some limitations to classifier accuracy arose from the dataset itself. The text contained opinions, descriptive sentences as well as quotations in indirect as well as direct speech. This was discussed with the annotators before and during the annotation process, as it was reported to have created some confusion. In the case of opinions, explicit expressions were easily distinguishable, but in many cases the opinions were implicit. Quotations were also problematic as these could easily be misinterpreted without the proper context that a paragraph might provide. Sarcasm and metaphoric speech is also among challenges that automatic classifiers have to face, e.g. "The protests were but shouts in the deserts because the wheel of racial equality had already been set on its way." (*Kuid protestid jäid hüüdjaks hääleks kõrbes, sest rassilise võrdsuse ratas oli juba hooga veerema lükatud.*). We also included keywords often used by the radical right to negatively refer to the liberals, like "multiculturalists" and "globalists" etc., which may be difficult to spot as negative without context or prior knowledge. Annotators also reported pro immigration stances as harder to identify. This may be due to anti-immigration rhetoric being more systematic and less fragmented whilst pro-immigration rhetoric is more dependent on the specific sub-topic.

**Exemplary analysis**

Lastly we conducted an exemplary diachronic analysis of the change of stances towards immigration across time. This tests the applicability of our method and demonstrates some of its possible uses. In the following, we visualize and analyze the larger changes in the stance trends in relation to media events, look at the related media polarization and general similarities based on text embeddings.

The relative amount of immigration related articles across time and publisher (see Fig. 3) provides an understanding of immigration related media events and their importance for each of the publishers. Uued Uudised clearly focuses more on the immigration topic than Ekspress Grupp, based on keyword prevalence. Uued Uudised also has a stronger reaction to immigration related media events, such as the European migration crisis of 2015-2016, UN immigration pact at the end of 2018, and the Russian invasion of Ukraine from February 2022 onwards, which caused an increase in refugees. These findings confirm what is known about radical-right media in general and it provides novel insight into the Estonian context.

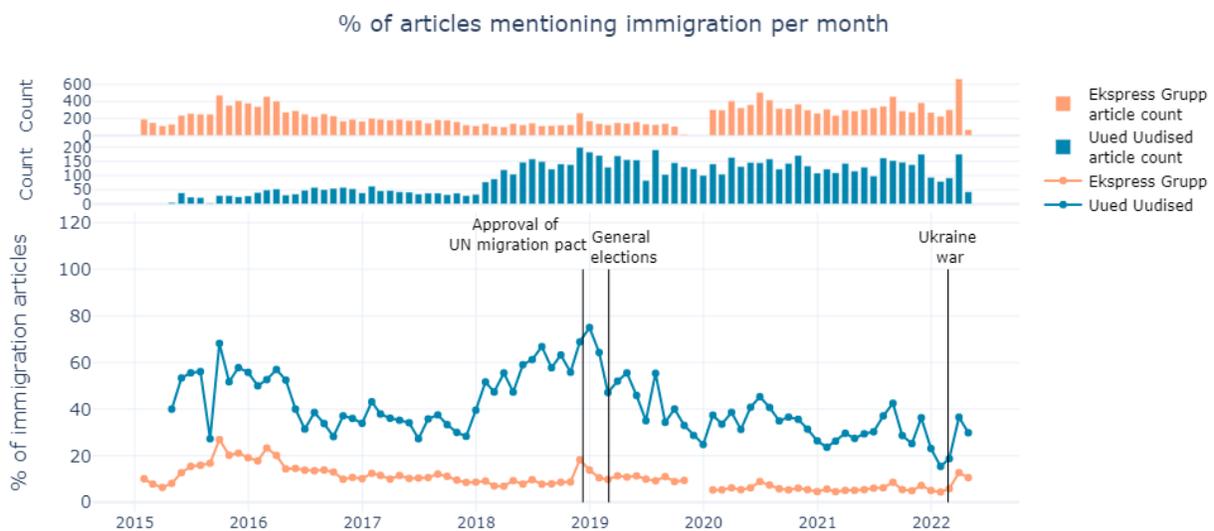

**Figure 3. Percentage of articles mentioning immigration**. Top plots show the counts of articles mentioning immigration. The articles contain at least one immigration related keyword. Higher percentage for the populist radical-right source (blue) confirms that the outlet is more focused on the immigration. The fluctuations in Uued Uudised is due to the smaller amount of data in absolute terms, especially in 2015. The relatively lower amount of immigration related articles in Ekspress group data since 2020 is likely connected to the significantly increased amount of content from a larger variety of specific journals, indicating that the amount of immigration related content is somewhat dependent on specific journals of Ekspress Group (see appendix on Ekspress Grupp distribution details).

We used the best-performing model, based on Est-RoBERTa, to predict the stances of all sentences in the corpus containing relevant keywords (n=106539). We focus on monthly trends, as a tradeoff between detail and the amount of available data per unit of time.

The findings, as seen in Figs. 4 and 5, confirm and expands previous assumptions and findings within media studies on the roles of the respective publishers (Kasekamp et al., 2019) and radical right populism in general (cf. (Mudde, 2007; Rooduijn et al., 2014). We found trends that showed polarization and indicated changes of stance corresponding to the UN migration pact and elections, and Ukraine war. Uued Uudised stance was generally against immigration, not neutral or supportive. On the other hand, Ekspress Grupp had a dominantly neutral stance over time and kept generally more stable than Uued Uudised. The relative stance differed noticeably per keyword group, whereas multiculturalism and xenophobia and race related words had the highest percentage of sentences labeled as Against migration (cf. Figs. S6-S9 for stances per keyword groups).

There is a clear change taking place around the 2018-2019 during the UN migration pact discussions (most heated debates in Estonian media happening around November 2018) and general elections (March 3, 2019). Uued Uudised contained more sentences classified as Against migrants than before and right after that period. The share of the Against stance is increasing with the UN migration pact discussions, but decreases soon after the elections in March 2019. The Against stance increased in these years for all of the keyword groups. A change is also noticeable in Ekspress Grupp, where relevance of Against stance increases during the same period. This demonstrates the possible connection between potential politicization of the migration topic and the elections. This could be further investigated in future research.

From March 2020, when Covid-19 became the dominant media event, the stances seem to change again. This may be due to the shift of focus on other topics, such as Covid, where the radical-right shifted its focus from anti-immigration to anti-governmental. Lastly, the events of the Russian invasion into Ukraine in 2022 correspond to a small increase in supportive stance in Uued Uudised and a much larger increase in supportive stance towards immigrants in Ekspress Grupp. Whilst the Supportive stance increased in almost all of the keyword groups for Ekspress Grupp, there was more variability for Uued Uudised. This ambiguity of Uued Uudised may reflect the continued anti-immigration rhetoric by the related rightwing political party of EKRE.

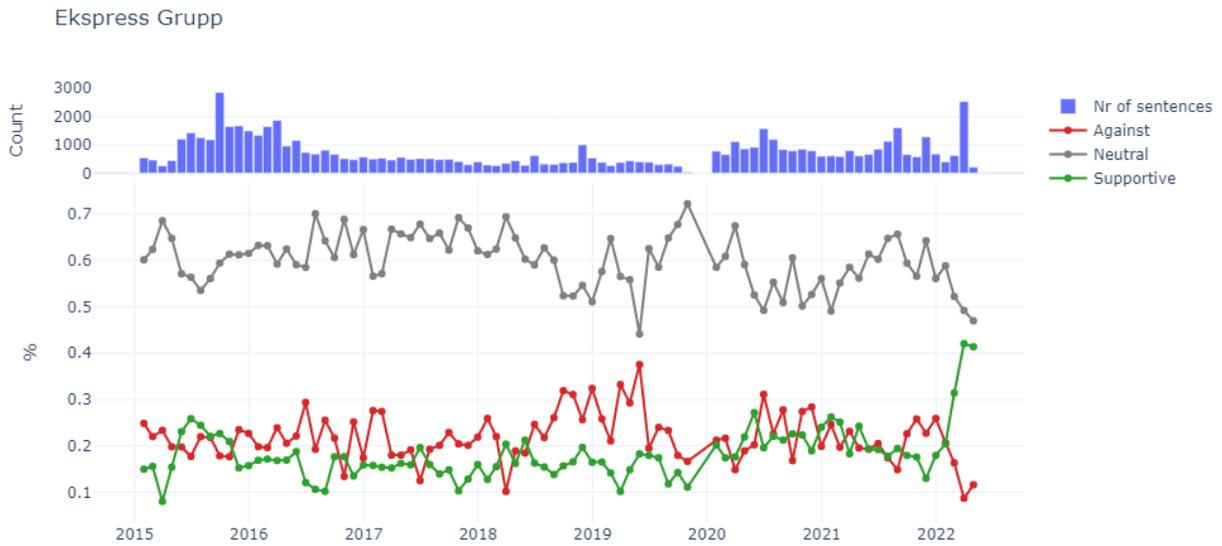

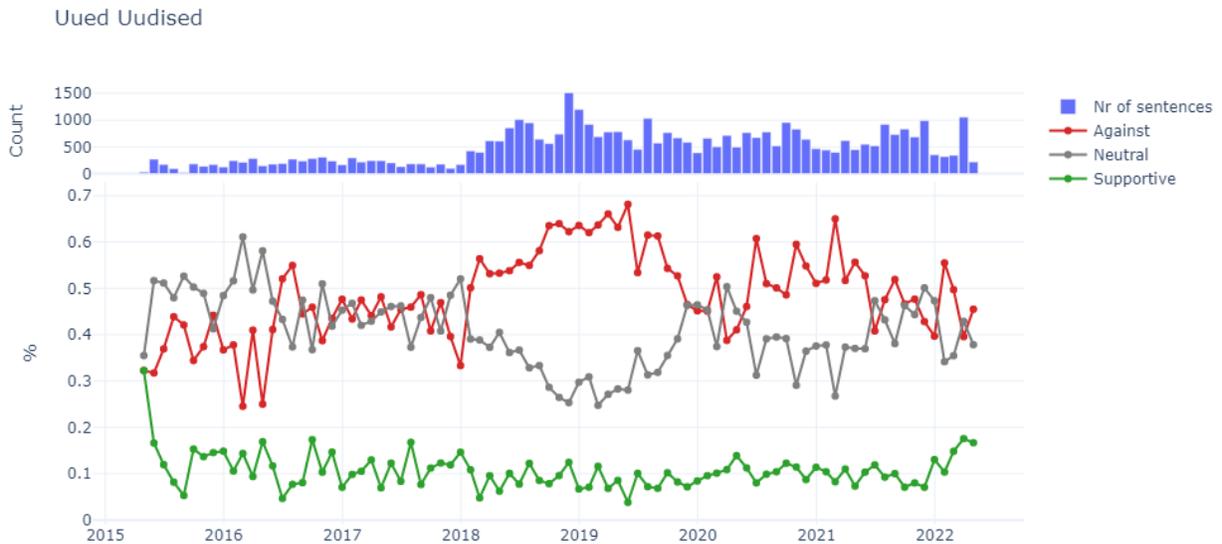

**Figures 4 and 5. Stances of immigration related sentences**. It shows the relative percentage of each stance per month for both publishers. Barplots show the amount of immigration sentences per month in comparison. In 2022 at the beginning of the Ukraine war, there was a noticeable increase in Supportive stances towards migration in the Ekspress Group with a much smaller increase in Uued Uudised (cf. Figs. S10 and S11 for trends with thresholds for classification certainty).

In order to understand the changes taking place within and between the publishers, we calculated the cosine similarities between the sentences from different publishers with Sentence-BERT (Reimers & Gurevych, 2020). Figure 6 shows how the cosine similarity has spiked in the end of 2018 and 2022. We interpret the change as a possible increase of similarities of topics or rhetorics towards immigration. The latter change with the Ukrainian war differs from one connected to the 2018 UN migration pact as the similarity increased almost just as much with all of the stances. Analysis of similarities within publisher sources (cf. Fig. S12) had similar trends relating to those events, meaning that both publishers were possibly more focused on one media or used similar rhetoric during these months.

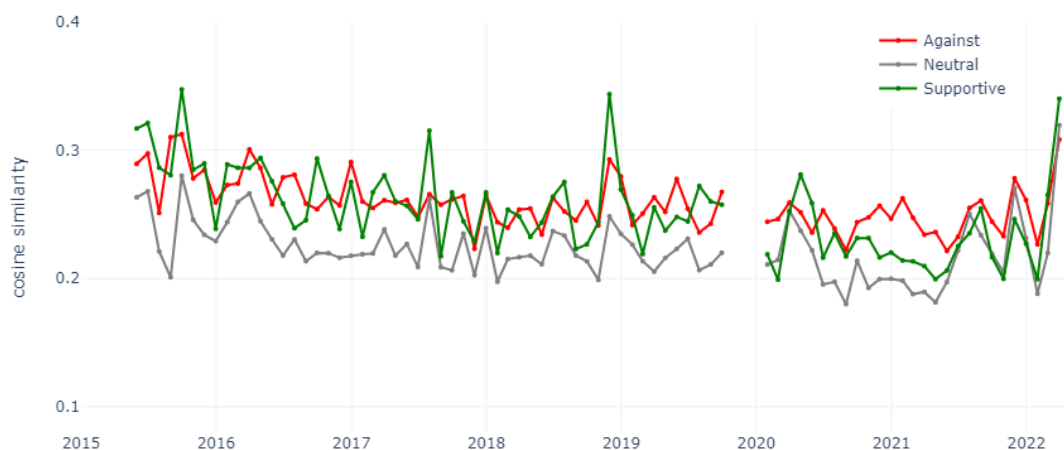

**Fig 6. Comparison on sentence cosine similarities between the publishers.** Similarities are calculated separately per stance. The larger spikes in higher cosine similarities in the end of 2018 and in 2022 may be indicating that the outlets pay attention to similar events in a similar stance.

## Discussion

Our study shows that automated stance detection is feasible and provides useful insights for media monitoring and analytics purposes, also beyond large languages like English or German. The accuracy of the classifiers was satisfactory, achieving F1 macro 0.66 with Est-RoBERTa. Zero-shot ChatGPT achieved a similar result of 0.65. We expect zero-shot accuracy will improve as generative AI models are being improved and developed. Classification of Against stances was noticeably more accurate than of Supportive stances. As expected, radical-right news media indeed holds generally a more anti-immigration stance in comparison to more mainstream news. We also provided insights into stance change over time, relating it to known local and world events, identifying increased interest towards these topics during the 2015-2016 immigration crisis, the 2018-2019 UN immigration pact and local elections, and the 2022 Russian invasion into Ukraine. These findings, approximated by applying an automated classifier, can be used as basis for further more in-depth research in Estonian-specific or areal media and politics studies.

However, there are also limitations. Fine-tuning pretrained LLMs as classifiers requires annotated training data, which may not be available for specific topics or in lower-resource languages. We discussed issues with annotation, pointing out that linguistically and socio-politically complex topics such as this are also difficult for human annotators and for formalizing the task. There is also the question of unit of analysis: shorter units like sentences are fast to annotate, but may not contain enough contextual information. Longer like paragraphs do, but may contain multiple stances, which complicates the task both for humans and machines.

**Future research**

Whilst supervised stance detection can provide acceptable results, the need for annotated training data makes it time consuming and expensive, while being applicable to one topic at a time. One option is to use a generic sentiment classifier instead. However, we showed that this does not work very well for complex topics such as immigration, where support may be expressed in sentences with negative overall tone, and vice versa. Using new generation generative LLMs like ChatGPT may provide a solution, being easy to instruct in natural language, and applicable across languages, tasks and topics. This makes it particularly attractive for smaller languages with less resources and with less existing annotated datasets.

These models could also be used to annotate data in tandem with human annotators, or augment existing annotations (Gilardi et al., 2023). Accuracy and model bias should still be evaluated. For example, in our case it could have been used to further classify sentences as expressing opinions, factual descriptions, and direct quotes. This can result in a feedback loop that results in better datasets, more accurate models and also better understanding of the functioning of the model through assessing the classification errors.

This new approach has already been explored in preliminary experiments, which our research complements. The accuracy of applying zero-shot learning should still be evaluated, and not be taken for granted. For this, annotated datasets such as the one we also make available, are still useful and necessary. This is more so relevant in smaller languages, for which likely less initial training data has been used in models like ChatGPT. While annotating new datasets requires instructing human annotators, using generative models requires careful prompt engineering. This also complicates replication of results, as slightly different instructions can lead to differences in classification performance, in addition to the inherently stochastic nature of generative AI (Reiss, 2023). More so, results are difficult to replicate if a cloud-based, frequently updated model like ChatGPT is used. Then again, these models may also improve accessibility, making deep learning available and feasible for non-computer scientists and researchers with limited access to large computer clusters, although it will depend on the companies offering such services and their pricing policies. Therefore, we expect generative AI driven analytics to become more widespread together with the affordances of cloud based computing across disciplines. This also calls for more critical studies as well as thorough analysis of the applications of the methods to better understand the biases related to specific LLMs and cloud-based services.

## Conclusions

We demonstrated the applicability of automated stance detection using pretrained LLMs for socio-politically complex topics in smaller languages on the example of Estonian news media coverage of immigration discourse. We compare several popular models, and also release the stance-annotated dataset. Our experiments with using ChatGPT as an instructable zero-shot classifier are promising, and if applied carefully, this approach could obviate the need for topic-specific annotation and expedite media analytics and monitoring tasks. This is more so the case in languages where such resources are limited. As a proof of concept, we also applied one classifier to the larger corpus to provide an overview of changes in immigration in Estonian news media in 2015-2022, including one mainstream and one radical-right news source, finding support for discussions in previous literature as well as providing new insights.

## Declarations

Annotation and data acquisition together with preliminary analysis were conducted with co-funding from Ekspress Grupp, which did not influence the design of the study nor the conclusions. M.M., A.K., M.S., I.I. are supported by the CUDAN ERA Chair project for Cultural Data Analytics at Tallinn University, funded through the European Union Horizon 2020 research and innovation program (Project No. 810961).

## Data and code availability

Data and code used in this study are open access and available in this GitHub repository: https://github.com/markmets/immigration-prediction-EST

## Author contributions

Mark Mets contributed Conceptualization, Data Curation, Investigation, Software, Formal analysis, Methodology, Visualization, and Writing – original draft. Andres Karjus contributed Conceptualization, Methodology, Supervision, and Writing – original draft. Indrek Ibrus contributed Conceptualization, Methodology, Supervision, Writing – review & editing. Maximilian Schich contributed Conceptualization, Supervision and Writing – review & editing.

## References

Abts, K., & Rummens, S. (2007). Populism versus Democracy. *Political Studies*, *55*(2), 405–424. https://doi.org/10.1111/j.1467-9248.2007.00657.x

Aiyappa, R., An, J., Kwak, H., & Ahn, Y.-Y. (2023). Can we trust the evaluation on ChatGPT? *ArXiv:2303.12767 [Cs]*. Accessed 10 May 2023.

ALDayel, A., & Magdy, W. (2021). Stance detection on social media: State of the art and trends. *Information Processing & Management*, *58*(4), 102597. https://doi.org/10.1016/j.ipm.2021.102597

Allaway, E., & McKeown, K. (2020). Zero-Shot Stance Detection: A Dataset and Model using Generalized Topic Representations. *Proceedings of the 2020 Conference on Empirical Methods in Natural Language Processing*, 8913–8931.

Auers, D. (2018). Populism and Political Party Institutionalisation in the Three Baltic States of Estonia, Latvia and Lithuania. *Fudan Journal of the Humanities and Social Sciences*, *11*(3), 341–355. https://doi.org/10.1007/s40647-018-0231-1

Batanović, V., Cvetanović, M., & Nikolić, B. (2020). A versatile framework for resource-limited sentiment articulation, annotation, and analysis of short texts. *PLOS ONE*, *15*(11), e0242050. https://doi.org/10.1371/journal.pone.0242050

Bender, E. M., Gebru, T., McMillan-Major, A., & Shmitchell, S. (2021). On the Dangers of Stochastic Parrots: Can Language Models Be Too Big? 🦜. *Proceedings of the 2021 ACM Conference on Fairness, Accountability, and Transparency*, 610–623.

https://doi.org/10.1145/3442188.3445922

Braghiroli, S., & Makarychev, A. (2023). Conservative populism in Italy and Estonia: Playing the multicultural card and engaging "domestic others." *East European Politics*, *39*(1), 128–149. https://doi.org/10.1080/21599165.2022.2077725

Burscher, B., van Spanje, J., & de Vreese, C. H. (2015). Owning the issues of crime and immigration: The relation between immigration and crime news and anti-immigrant voting in 11 countries. *Electoral Studies*, *38*, 59–69. https://doi.org/10.1016/j.electstud.2015.03.001

Card, D., Chang, S., Becker, C., Mendelsohn, J., Voigt, R., Boustan, L., Abramitzky, R., & Jurafsky, D. (2022). Computational analysis of 140 years of US political speeches reveals more positive but increasingly polarized framing of immigration. *Proceedings of the National Academy of Sciences*, *119*(31), e2120510119. https://doi.org/10.1073/pnas.2120510119

Conneau, A., Khandelwal, K., Goyal, N., Chaudhary, V., Wenzek, G., Guzmán, F., Grave, E., Ott, M., Zettlemoyer, L., & Stoyanov, V. (2020). Unsupervised Cross-lingual Representation Learning at Scale. *Proceedings of the 58th Annual Meeting of the Association for Computational Linguistics*, 8440–8451. https://doi.org/10.18653/v1/2020.acl-main.747

Devlin, J., Chang, M.-W., Lee, K., & Toutanova, K. (2019). BERT: Pre-training of Deep Bidirectional Transformers for Language Understanding. *Proceedings of the 2019 Conference of the North American Chapter of the Association for Computational Linguistics: Human Language Technologies*, *Volume 1*, 4171–4186.

Do, H. H., Prasad, P., Maag, A., & Alsadoon, A. (2019). Deep Learning for Aspect-Based Sentiment Analysis: A Comparative Review. *Expert Systems with Applications*, *118*, 272–299. https://doi.org/10.1016/j.eswa.2018.10.003

Du, J., Xu, R., He, Y., & Gui, L. (2017). Stance Classification with Target-specific Neural Attention. *Proceedings of the 2017 Conference on Empirical Methods in Natural Language Processing*, 3988–3994.

Engesser, S., Ernst, N., Esser, F., & Büchel, F. (2017). Populism and social media: How politicians spread a fragmented ideology. *Information, Communication & Society*, *20*(8), 1109–1126. https://doi.org/10.1080/1369118X.2016.1207697

Ghosh, S., Singhania, P., Singh, S., Rudra, K., & Ghosh, S. (2019). Stance Detection in Web and Social Media: A Comparative Study. In F. Crestani, M. Braschler, J. Savoy, A. Rauber, H. Müller, D. E. Losada, G. Heinatz Bürki, L. Cappellato, & N. Ferro (Eds.), *Experimental IR Meets Multilinguality, Multimodality, and Interaction*

(pp. 75–87). Springer International Publishing. https://doi.org/10.1007/978-3-030-28577-7_4

Gilardi, F., Alizadeh, M., & Kubli, M. (2023). ChatGPT Outperforms Crowd-Workers for Text-Annotation Tasks. *ArXiv:2303.15056 [Cs]*. Accessed 10 Mai 2023.

Hedderich, M. A., Lange, L., Adel, H., Strötgen, J., & Klakow, D. (2021). A Survey on Recent Approaches for Natural Language Processing in Low-Resource Scenarios. *Proceedings of the 2021 Conference of the North American Chapter of the Association for Computational Linguistics: Human Language Technologies*, 2545–2568. https://doi.org/10.18653/v1/2021.naacl-main.201

Kaal, E., & Renser, B. (2019). *Rändetemaatika kajastamine Eesti meedias*. MTÜ Mondo. https://mondo.org.ee/wp-content/uploads/2019/09/R%C3%A4ndetemaatika-kajastamine-Eesti-meedias.pdf

Kasekamp, A., Madisson, M.-L., & Wierenga, L. (2019). Discursive Opportunities for the Estonian Populist Radical Right in a Digital Society. *Problems of Post-Communism*, *66*(1), 47–58. https://doi.org/10.1080/10758216.2018.1445973

Khatua, A., & Nejdl, W. (2022). Unraveling Social Perceptions & Behaviors towards Migrants on Twitter. *Proceedings of the Sixteenth International AAAI Conference on Web and Social Media (ICWSM 2022)*, *16*, 512–523.

Koppel, K., & Jakobson, M.-L. (2023). Who Is the Worst Migrant? Migrant Hierarchies in Populist Radical-Right Rhetoric in Estonia. In M.-L. Jakobson, R. King, L. Moroşanu, & R. Vetik (Eds.), *Anxieties of Migration and Integration in Turbulent Times* (pp. 225–241). Springer International Publishing. https://doi.org/10.1007/978-3-031-23996-0_13

Küçük, D., & Can, F. (2020). Stance Detection: A Survey. *ACM Computing Surveys*, *53*(1), 1–37. https://doi.org/10.1145/3369026

Li, W., Gao, S., Zhou, H., Huang, Z., Zhang, K., & Li, W. (2019). The Automatic Text Classification Method Based on BERT and Feature Union. *2019 IEEE 25th International Conference on Parallel and Distributed Systems (ICPADS)*, 774–777. https://doi.org/10.1109/ICPADS47876.2019.00114

Liu, P., Yuan, W., Fu, J., Jiang, Z., Hayashi, H., & Neubig, G. (2023). Pre-train, Prompt, and Predict: A Systematic Survey of Prompting Methods in Natural Language Processing. *ACM Computing Surveys*, *55*(9), 1–35. https://doi.org/10.1145/3560815

Lotman, Yu. M., Uspensky, B. A., & Mihaychuk, G. (1978). On the Semiotic Mechanism of Culture. *New Literary History*, *9*(2), 211. https://doi.org/10.2307/468571

Madisson, M.-L., & Ventsel, A. (2016). Autocommunicative meaning-making in online communication of the Estonian extreme right. *Sign Systems Studies*, *44*(3), 326–354. https://doi.org/10.12697/SSS.2016.44.3.02

Madisson, M.-L., & Ventsel, A. (2018). Groupuscular identity-creation in online-communication of the Estonian extreme right. *Semiotica*, *2018*(222), 25–46. https://doi.org/10.1515/sem-2016-0077

Magueresse, A., Carles, V., & Heetderks, E. (2020). Low-resource Languages: A Review of Past Work and Future Challenges. *ArXiv:2006.07264 [Cs]*. Accessed 10 May 2023.

Meltzer, C. E., & Schemer, C. (2021). Miscounting the others: Media effects on perceptions of the immigrant population size. In *Media and Public Attitudes Toward Migration in Europe* (pp. 174–189). Routledge.

Mohammad, S. M., Sobhani, P., & Kiritchenko, S. (2017). Stance and Sentiment in Tweets. *ACM Transactions on Internet Technology*, *17*(3), 1–23. https://doi.org/10.1145/3003433

Mudde, C. (2007). *Populist radical right parties in Europe*. Cambridge University Press.

Pajupuu, H., Altrov, R., & Pajupuu, J. (2016). Identifying Polarity in Different Text Types. *Folklore: Electronic Journal of Folklore*, *64*, 125–142. https://doi.org/10.7592/FEJF2016.64.polarity

Pang, B., & Lee, L. (2008). Opinion Mining and Sentiment Analysis. *Foundations and Trends® in Information Retrieval*, *2*(1–2), 1–135. https://doi.org/10.1561/1500000011

Petsinis, V. (2019). Identity Politics and Right-Wing Populism in Estonia: The Case of EKRE. *Nationalism and Ethnic Politics*, *25*(2), 211–230. https://doi.org/10.1080/13537113.2019.1602374

Qin, C., Zhang, A., Zhang, Z., Chen, J., Yasunaga, M., & Yang, D. (2023). Is ChatGPT a General-Purpose Natural Language Processing Task Solver? *ArXiv:2302.06476 [Cs]*. Accessed 10 May 2023.

Reimers, N., & Gurevych, I. (2020). Making Monolingual Sentence Embeddings Multilingual using Knowledge Distillation. *Proceedings of the 2020 Conference on Empirical Methods in Natural Language Processing*, 4512–4525.

Reiss, M. V. (2023). Testing the Reliability of ChatGPT for Text Annotation and Classification: A Cautionary Remark. *ArXiv:2304.11085 [Cs]*. Accessed 10 May 2023.

Rooduijn, M., de Lange, S. L., & van der Brug, W. (2014). A populist *Zeitgeist*? Programmatic contagion by populist parties in Western Europe. *Party Politics*, *20*(4), 563–575. https://doi.org/10.1177/1354068811436065

Saarts, T., Jakobson, M.-L., & Kalev, L. (2021). When a Right-Wing Populist Party Inherits a Mass Party Organisation: The Case of EKRE. *Politics and Governance*, *9*(4), Article 4. https://doi.org/10.17645/pag.v9i4.4566

Sahlgren, M. (2008). The distributional hypothesis. *Italian journal of linguistics, 20*(1), 33-54.

Sobhani, P., Inkpen, D., & Zhu, X. (2017). A Dataset for Multi-Target Stance Detection. *Proceedings of the 15th Conference of the European Chapter of the Association for Computational Linguistics: Volume 2, Short Papers*, 551–557. https://aclanthology.org/E17-2088

Tanvir, H., Kittask, C., Eiche, S., & Sirts, K. (2021). EstBERT: A Pretrained Language-Specific BERT for Estonian. *Proceedings of the 23rd Nordic Conference on Computational Linguistics (NoDaLiDa)*. 11–19

Ulčar, M., Žagar, A., Armendariz, C. S., Repar, A., Pollak, S., Purver, M., & Robnik-Šikonja, M. (2021). Evaluation of contextual embeddings on less-resourced languages. *ArXiv:2107.10614 [Cs]*. Accessed 10 Mai 2023.

Vamvas, J., & Sennrich, R. (2020). X-Stance: A Multilingual Multi-Target Dataset for Stance Detection. *5th Swiss Text Analytics Conference (SwissText) & 16th Conference on Natural Language Processing (KONVENS)*.

Yantseva, V., & Kucher, K. (2021). Machine Learning for Social Sciences: Stance Classification of User Messages on a Migrant-Critical Discussion Forum. *2021 Swedish Workshop on Data Science (SweDS)*, 1–8. https://doi.org/10.1109/SweDS53855.2021.9637718

Zhang, B., Ding, D., & Jing, L. (2023). How would Stance Detection Techniques Evolve after the Launch of ChatGPT? *ArXiv:2212.14548 [Cs]*. Accessed 10 Mai 2023.
24

# APPENDIX

1. **Keywords details**

Regex search that looked for different forms of words whilst avoiding non topical but form-wise similar words. These are the regex searches divided into eight topics in the original language and a translation of examples to provide an idea of which keywords it applies to.

**Migration (*Ränne*) - captures keywords like "migration" and "migrant"**
'migrats|migran|migreer|ränne|rände|rännet|sisserända|sisse rända|väljarända|töörän[dn]|õpirän[nd]|pendelrän[nd]|hiigelrän[nd]|tagasirän[dn]|massirän[dn]|rändle|rännelnud|tagasipöördu[vj]ad|paadirän[dn]|väljarän[nd]|massipagem|legaal[ns]e[t]*rän[nd]|edasirän[nd]|seaduslik[kult]*
rän[nd]|ringirända[vj]|rändajate ümberpaigu|juhitav[a]* rän[nd]|rända[sivad]* sisse|sisse- ja läbiränd|rändajatemass|kodutud ja rändajad|rahvasterän[nd]|rahvaste rän[nd]'
+ additional filter removes mostly animal migration and migraine related sentences:
'lind|linnu|lindu|kala|loom|imetaja|migreen|kahepaik|roomaja|hani|hane|ogavalk|relve|kuula rändajat'

**Refugees (*pagulased*) - captures keywords like "refugee", "asyluym seeker", "illegal (immigrant)" and "border control"**
'pagula|asüül|varjupaigataotl|põgenik|inimkaub|illegaal|piirikontroll '

**Foreign workers (*välistööjõud*) - captures keywords like "foreign workers", "(digital) nomad"**
'välistööjõu|tööjõu sisse|võõrtöö|hooajatöö|välismaala|võõramaala|nomaad'

**Foreign students (*välistudengid*) - captures keywords like "foreign student"**
'Välistuden|välisüliõpila'

**Noncitizens (*mittekodanikud*) - captures keywords like "living permit", "non-Estonian", "Estonian visa"**
'mittekodanikud': 'elamisloa|elamisluba|eesti viisa|viibimisalus|mitte-eestla|muula',

**Radical-right vs liberal opposition (*paremäärmus*) - captures keywords like "globalism", "new-europeans", "open borders", xenophobe", "multicultural"**

'globalist|globalism|uuseuroopla|suur asendami|suure asendamise|avatud uste poliitika|avatud piir|ksenofoob|võõrahirm|multikult',

**Race (*rass*) - captures keywords related to black and darker races and asians**
'([\W ]|^)neeg|mustanahali|([\W ]|^)rass|must, näita ust|europiid|negriid|mongoliid|asiaa|tõmmu|murjam|must mees|mustad mehed|mustad inimesed|must inimene',

**Ethnicity (*rahvus*) - captures keywords related to more often migration related ethnicities and locals, like "african", "moslem", "islamic", "arabic", "syrian", "vietnamese", "afghan", "iraqi", "sudanese", "ukrainian construction worker"**
'aafrikla|moslem|islam|araabla|süürla|vietnamla|afgaan|iraakla|iraanl a|sudaanla|ukraina ehitaja'

## 2. Data distribution

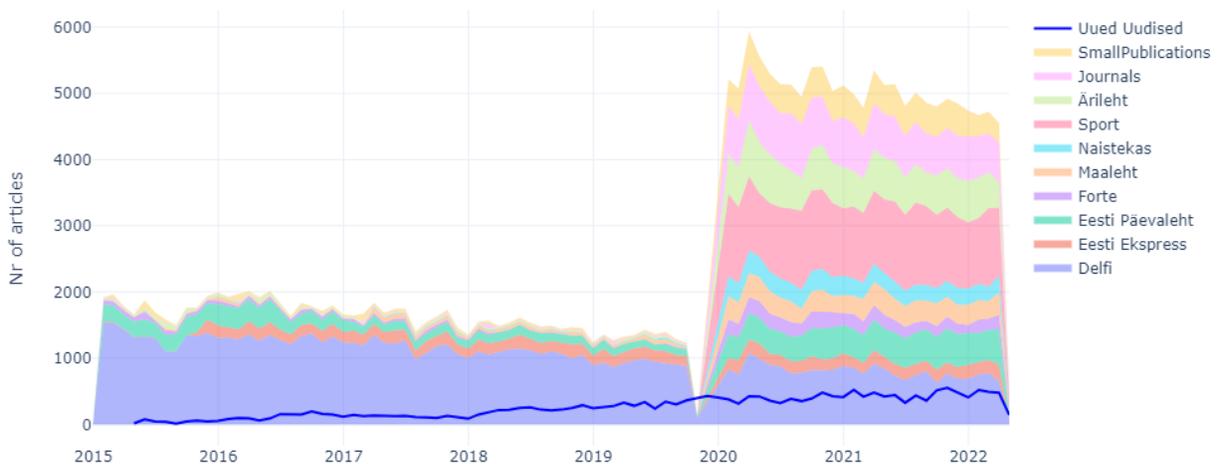

**Fig S1. Distribution of all articles in our dataset per periodical**. Area chart distinguishes the biggest periodicals of Ekspress Grupp per month and the blue trend compares it to the Uued Uudised. All Ekspress Grupp articles from 2015 to 2020 data mostly originates from Delfi, a fully

online platform. From 2020 onwards, the data contains almost all of the other biggest papers as well, including daily newspaper Eesti Päevaleht, weekly news Eesti Ekspress, a Sports portal, Maaleht focusing on rural topics, Ärileht on business and Forte on Technology. The graph excludes numerous other smaller publications that make up a relatively small amount of data. The difference comes from the fact that this was the data given by Ekspress Grupp.

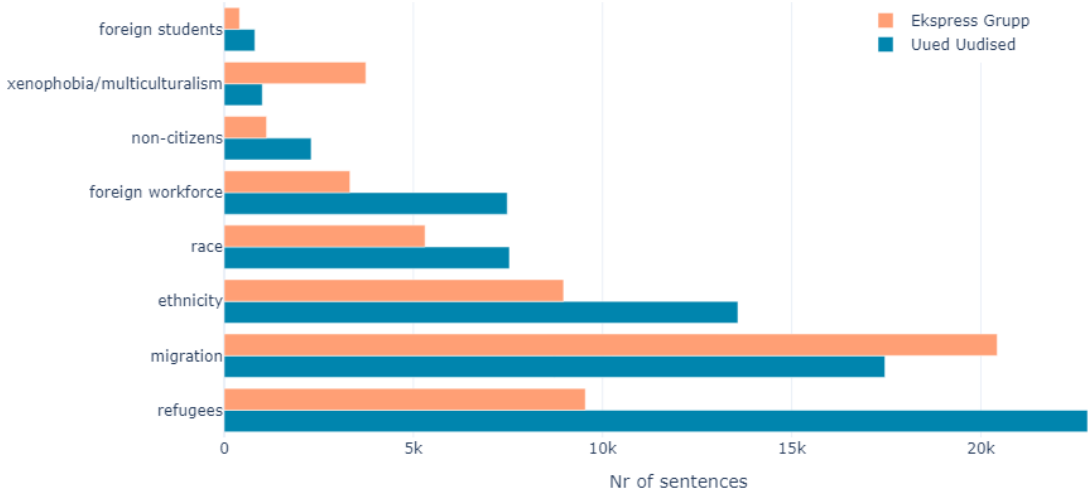

**Figure S2. Distribution of keyword groups by the number of sentences mentioning keywords relevant to the group**. The refugee and migration related keywords make up most of the dataset whilst there are relatively very few sentences about foreign students. The two outlets have some differences. E.g. Uued Uudised has double the amount of sentences on refugee and foreign workforce topics. Thirdly, xenophobia and multiculturalism related keywords are more used in Ekspress Grupp although there are differences in specific keywords. This indicates that these publishers have different focus on immigration related topics.

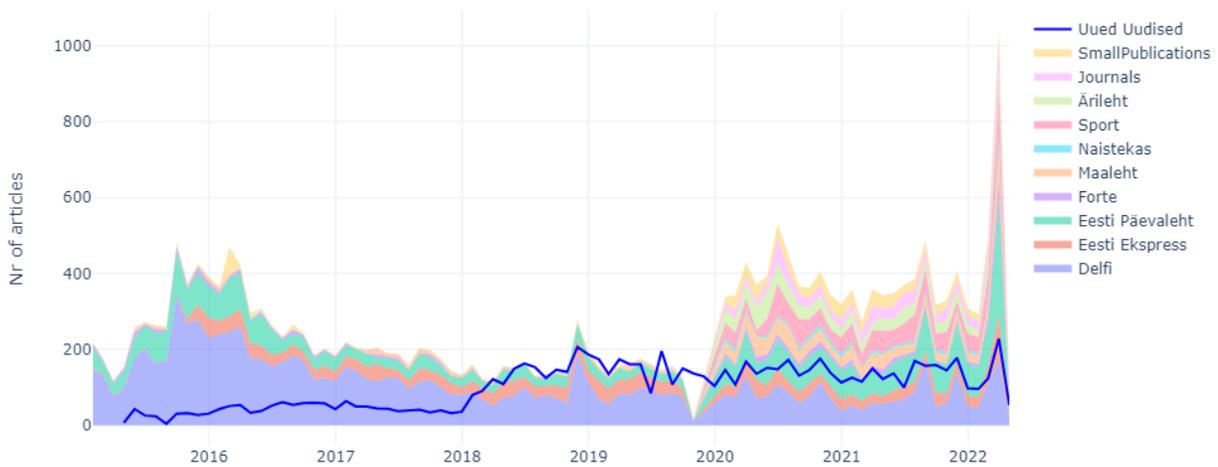

**Figure S3. Distribution of immigration related articles per periodical.** Shows the articles with immigration related keywords per largest periodicals in Ekspress Grupp in our dataset. Stacked colors represent different publications in Ekspress Grupp. The main source is the online platform Delfi (purplish). Compared to Uued Uudised marked with blue line.

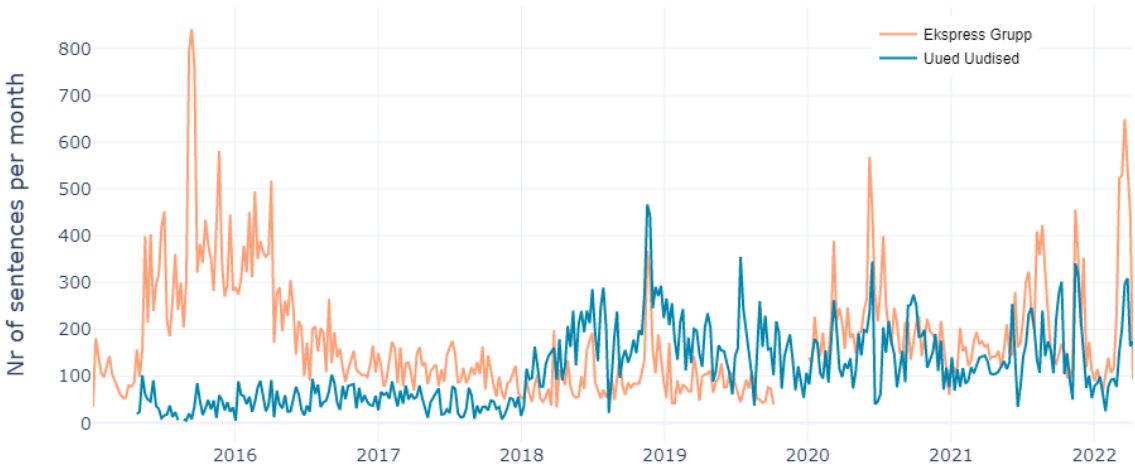

**Figure S4. Weekly distribution of immigration related sentences**. A more detailed view of immigration related events on a weekly scale. Provided as a comparison to monthly trends

which shows that dependence of monthly trends are dependent on more specific weekly changes.

## 3. Explanations on LLM and automatic classifier

**Text embeddings explanation**

We provide a short explanation of LLMs and the process of classification. LLMs are based on text embeddings. It means that the text is first transformed into numerical form based on the text co-occurrences, theoretically based on the distributional hypothesis, stating that a meaning of a word can be inferred from the lexical context, that is, from the words around it (Sahlgren, 2008). Therefore, the goal of the model is to predict the probabilities of words occurring amongst other words. The same logic applies to ChatGPT. Next, the text undergoes dimension reduction and is represented in a multidimensional space where the number of dimensions is dependent on a specific model (e.g. 768 for BERT). Every dimension represents a certain abstract feature of the word, like the "blueness" or "wealth" although in practice these dimensions are hard to identify. The resulting embedding space affords us to rely on a spacial interpretation of textual relations, e.g. how far are terms "immigration" and "refugees" from each other in relation to other terms, or the comparison of distances between whole sentences as in our case.

**Classification explanation**

The classification model uses the given parameters to figure out what sentence corresponds to which category through trial and error of labeling the training data. The approach uses a process called masking where the model either tries to guess which word fits in the empty slot around other words or which words could be present in the empty slot around a specific one. The resulting model is expected to be generalizable for other similar data, measured with the part of data that is kept back from the annotations to test its performance on data not used for training. BERT has been reported to often outperform large language models, like Word2Vec or GloVe (Devlin et al., 2019), although a more complicated combination of BERT with other methods may yield even somewhat better results (Li et al., 2019), not to mention larger models, like the GPTs. BERT follows models like Word2Vec, Glove and ELMo and is superseded by similar but even larger language models, like RoBERTa. All of them use pre-trained language models, meaning that they are pre-trained on large language corporas. Typically these are further fine-tuned, like in our case, to fit a more specific task. They differ in the size of this pre-training language model, but all use datasets of millions of words. What sets them apart is that Word2Vec and Glove give only one vector per word, but ELMo and BERT are context sensitive. Meaning that the word *bank* in phrase *river bank erodes* gets a separate numerical value than *bank* in phrase *financial bank* *refinances.* In comparison to ELMo, BERT is better at this by simultaneously considering both the text that comes before and after the word *bank*.

## 3. Examples of stance annotations

| Stance class | Nr of sentences | Example (translated) | Example (Original) |
|---|---|---|---|
| Against (1-2) | n=1149 | Massimmigration would be disastrous for Europe and it would not solve anything in the world.<br><br>Migrants who are at least somewhat suspicious must be expelled from Estonia in the interests of security. | Massiimmigratsioon oleks Euroopale hukatuslik ja see ei lahendaks maailmas mitte midagi.<br><br>Vähegi kahtlased siin viibivad migrandid tuleb turvalisuse huvides Eestist välja saata. |
| Neutral (3) | n=1565 | Democrats blame the administration for using pandemic against immigration.<br><br>70% of the company's employees are foreigners. | Demokraadid süüdistavad administratsiooni pandeemia kasutamises sisserände vastu.<br><br>70% ettevõtte töötajatest on välismaalased. |
| Pro (4-5) | n=484 | The process to get a residence permit here was not very complicated.<br><br>The Spanish government announced on Tuesday that it would simplify rules for migrants and the unemployed to get jobs in agriculture during the coronavirus pandemic. | Protsess, et saada siin elamisluba, ei olnud väga keeruline.<br><br>Hispaania valitsus teatas teisipäeval, et lihtsustab reegleid migrantidele ja töötutele põllumajanduses töö saamiseks koroonaviiruse pandeemia ajal. |
| Ambiguous<br><br>(several viewpoints mixed together OR hard to say OR non-related keyword) | n=4117 | One can only wonder - when do Libyans quit and follow the flow of things when Europe is just talking about controlling the migrant crisis but itself just pours oil on fire.<br><br>In the past, Kristiina suffered from frequent migraines, which could paralyze a woman and force her into bed for the day, where the only way to cope was to keep a blanket over her head and her ears closed. | Jääb vaid küsida — millal ka liibüalased käega löövad ja toimuval vabavoolus minna lasevad, kui Euroopa vaid räägib rändekriisi ohjeldamisest, ise aga valab õli tulle?<br><br>Varasemalt vaevasid Kristiinat sagedased migreenid, mis võisid naist halvata ja sundida ta terveks päevaks voodisse, kus ainus võimalus hakkama saada oli hoida tekki pea peal ja kõrvasid kinni. |

**Table S1. Examples from our final stance training set representing each category.** Original punctuation.

## 4. Annotation interrater scores

|  | 6 categories (1-5, NA) n=550 | 4 categories (neg,neut,pos,mh) n=550 | 3 categories (neg,neut+mh,pos) n=550 | 3 categories (neg,neut,pos) n=222 | 2 categories (neg,pos) n=82 |
|---|---|---|---|---|---|
| Both Annotators | 0.45 | 0.49 | 0.57 | **0.68** | 0.97 |
| *Annotator-J* | 0.46 | 0.52 | 0.59 | **0.69** | 0.95 |
| *Annotator-N* | 0.45 | 0.45 | 0.54 | **0.66** | 1 |

**Table S2. Interrater agreement from a sample rated by third annotator (Cohen's kappa).**

## 5. Examples of sentiment annotations

| Class | Number of sentences | Example (translated) | Example (Original) |
|---|---|---|---|
| Negative (1-2) | n=1927 | The leaders of Estonian Air and the public should understand that the strategy does not work or does not work in the way that it is. In a simplified way, we could say that there are two options: to abolish the company or create a new conception | Estonian Airi juhtdel ning avalikkusel tuleks aru saada, et sellisel kujul strateegia ei toimi. Väga lihtsustatult võib väita, et võimalusi on kaks: firma likvideerida või luua täiesti uus kontseptsioon. |
| Neutral (3) | n=727 | Content wise it is the most complicated and delicate issue that could possibly rise in the doctor-patient relationship. | Sisult on tegemist kõige keerukama ja tundlikuma küsimusega, mis üldse võib patsiendi ja arsti suhetes tekkida. |
| Positive (4-5) | n=882 | He added "he is a very interesting person and his style and music are outstanding. I wish him good luck in America" | «Ta on väga huvitav inimene ning tema stiil ja muusika on väljapaistvad. Soovin talle Ameerikas edu,» lisas ta. |
| Contradictory | n=552 | The allies will come after that. Personal resistance does not warrant success – except in the fairytales – but it is still the only way to keep at least some kind of realistic hope for success. | Liitlased tulevad pärast seda. Isiklik vastuhakk ei garanteeri edu – välja arvatud muinasjuttudes –, kuid on siiski ainus viis, kuidas säilitada mingisugune reaalne edulootus. |

**Table S3. Examples from sentiment annotations from Pajupuu et al., 2016.**

## 6. Hyperparameters

Batch size: 16

Learning rate: 5e-5 (XLM-RoBERTa: 5e-6)

Epochs: 2 (XLM-RoBERTa: 5)

Warmup ratio: 0.1

## 7. Prediction mistake analysis examples

|  | **Human** | **AI** | **Probabilities** | **Example** (translated) | **Example** (original) |
|---|---|---|---|---|---|
| **Annotation problems** (doubtful annotation) | Pro<br><br>Pro | Against<br><br>Against | [0.7, 0.17, 0.13]<br><br>[0.64, 0.33, 0.04] | I would call on both sides - those who welcome the admission of refugees and those who fear it - not to give their voices to the silent ones.<br><br>President Macron wants to reshape relations between French Muslims and the secular French state. | Kutsuksin mõlemat poolt – nii pagulaste vastuvõtmise tervitajaid kui ka sellega hirmutajaid – mitte andma oma häält vaikijatele.<br><br>President Macron soovib ümber kujundada suhteid Prantsuse moslemite ja ilmaliku Prantsuse riigi vahel. |
| **Sarcasm** | Against<br><br>Pro | Pro<br><br>Against | [0.11, 0.36, 0.53] | Logically speaking, no ship in the Mediterranean should rescue a migrant ship that is sailing under its own power and not in imminent danger of sinking - bon voyage to Europe! | Loogiliselt võttes ei peaks ükski laev Vahemerel päästma migrandialust, mis liigub omal jõul ega ole otseses uppumisohus – head Euroopasse seilamist! |
| **Ambiguous and context dependent** (Meaning is context dependent, but human reader can infer it from the sentence) | Pro | Against | [0.85, 0.11, 0.04] | In Canada, a very pro-multiculturalist and pro-immigration Liberal Party is in power; it's leader and nation's prime minister Justin Trudeau has, for example, promised to raise his sons to be feminists | Kanadas on võimul igati multikultuursust ja sisserännet soodustav Liberaalne partei, mille liider ja riigi peaminister Justin Trudeau on lubanud näiteks oma poegadest feministid kasvatada. |
| | Pro | Against | [0.83, 0.11, 0.05] | The aim of the letter was to make the whole of society unanimously believe that Estonia has no future without Ukrainian migrant workers. | Kirja eesmärk oli panna kogu ühiskond üksmeelselt arvama, et Eesti riigil ilma Ukraina võõrtöölisteta pole tulevikku. |
| **Third person** (sentences talk about someone else's opinion. May often require contextual human knowledge) | Pro | Against | [0.75, 0.15, 0.1] | They criticize racism, homophobia, xenophobia and what they see as outdated nationalism. | Nad kritiseerivad rassismi, homovastasust, võõraviha ja nende arvates vananenud rahvuslust. |
| | Pro | Against | [0.58, 0.34, 0.08] | In a new Swedish version of "Lyckolandet" ("Happy Land"), Strömstedt expressed anti-racist views and named several people who are anti-immigrant. | Uute sõnadega rootsi keeles esitatud versioonis „Lyckolandet" („Õnnemaa") esitas Strömstedt rassismivastaseid seisukohti ja nimetas mitmeid sisserändevastaseid. |
| **Other mistakes** (sentences for which we could not ascertain causes of mistake) | Against | Pro | [0.05, 0.13, 0.83] | As an allied country, Estonia should also offer help in the forthcoming securing of the US southern border against the invasion of illegal immigrants, says Blue Dawn. | Liitlasriigina peaks Eesti pakkuma abi ka eesseisval USA lõunapiiri kindlustamisel illegaalsete immigrantide sissetungi vastu, leiab Sinine Äratus. |
| | Pro | Against | [0.87, 0.1, 0.04] | The tragic fate of refugees and their journey to Europe is the number one news story everywhere in the world, yet we have discovered that hate-mongers and provocateurs are disrupting intelligent debate rather than trying to find solutions. | Pagulaste traagiline saatus ja teekond Euroopasse on kõikjal maailmas uudis number 1, sellest hoolimata oleme avastanud, et vihakõnelejad ja provokaatorid segavad arukat debatti, mitte ei püüa leida lahendusi. |

**Table S4. Categorization of misclassified sentences.** Sentences wrongly classified by the model by mistaking Pro and Against classes. Examples are taken from shorter sentences. Original punctuation and spelling. The Probabilities column provides the classification probabilities for each example where numbers correspond in order to Against, Neutral and Supportive labels.

## 8. Detailed results of models

Based on 5x cross-validation of 20% of data

| Est-RoBERTa | | | |
|---|---|---|---|
| | f1-score | precision | recall |
| **Against** | 0.74 | 0.70 | 0.79 |
| **Neutral** | 0.55 | 0.50 | 0.63 |
| **Supportive** | 0.69 | 0.76 | 0.63 |
| **Micro avg** | 0.69 | 0.69 | 0.69 |
| macro avg | 0.66 | 0.65 | 0.68 |
| weighted avg | 0.69 | 0.70 | 0.69 |

| EstBert | | | |
|---|---|---|---|
| | f1-score | precision | recall |
| **Against** | 0.69 | 0.71 | 0.67 |
| **Neutral** | 0.53 | 0.51 | 0.56 |
| **Supportive** | 0.70 | 0.70 | 0.70 |
| **Micro avg** | 0.67 | 0.67 | 0.67 |
| macro avg | 0.64 | 0.64 | 0.64 |
| weighted avg | 0.67 | 0.67 | 0.67 |

| XLM-RoBERTa | | | |
|---|---|---|---|
| | f1-score | precision | recall |
| **Against** | 0.73 | 0.70 | 0.75 |
| **Neutral** | 0.54 | 0.45 | 0.69 |
| **Supportive** | 0.65 | 0.74 | 0.58 |
| **Micro avg** | 0.66 | 0.66 | 0.66 |
| macro avg | 0.64 | 0.63 | 0.67 |
| weighted avg | 0.66 | 0.68 | 0.66 |

| Multilingual BERT-UNCASED | | | |
|---|---|---|---|
| | f1-score | precision | recall |
| Against | 0.64 | 0.61 | 0.67 |
| Neutral | 0.38 | 0.32 | 0.48 |
| Supportive | 0.58 | 0.67 | 0.52 |
| Micro avg | 0.57 | 0.57 | 0.57 |
| macro avg | 0.54 | 0.53 | 0.56 |
| weighted avg | 0.57 | 0.60 | 0.57 |
| | | | |
| Multilingual BERT-CASED | | | |
| | f1-score | precision | recall |
| Against | 0.66 | 0.70 | 0.62 |
| Neutral | 0.40 | 0.34 | 0.48 |
| Supportive | 0.64 | 0.66 | 0.62 |
| Micro avg | 0.60 | 0.60 | 0.60 |
| macro avg | 0.56 | 0.57 | 0.58 |
| weighted avg | 0.61 | 0.62 | 0.60 |
| | | | |
| ChatGPT | | | |
| | f1-score | precision | recall |
| Against | 0.74 | 0.87 | 0.64 |
| Neutral | 0.64 | 0.56 | 0.75 |
| Supportive | 0.57 | 0.56 | 0.58 |
| Micro avg | 0.67 | | |
| macro avg | 0.65 | 0.66 | 0.66 |
| weighted avg | 0.67 | 0.67 | 0.69 |
| | | | |
| Sentiment (Est-RoBERTa) | | | |
| | f1-score | precision | recall |
| Against | 0.63 | 0.49 | 0.87 |
| Neutral | 0.42 | 0.70 | 0.30 |
| Supportive | 0.42 | 0.41 | 0.42 |
| Micro avg | 0.52 | | |
| macro avg | 0.49 | 0.53 | 0.53 |
| weighted avg | 0.49 | 0.58 | |
| | | | |
| | | | |

**Table S5. Detailed classification results per model.**

## 9. Confusion matrix of sentiment analysis

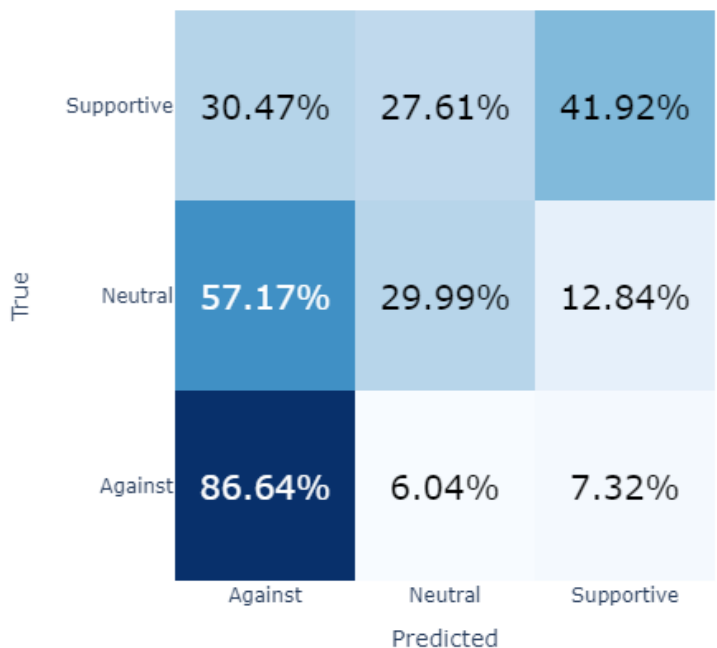

**Figure S5**. **Confusion matrix of model based on sentiment analysis dataset**. Compared to stance detection, there are noticeable more mistakes between supportive and Against classes and misclassification of neutral. On the other hand, a larger percentage of sentences was correctly classified as negative.

## 10. Stances of keyword groups

**Stance per keyword group**

In order to better understand the changes, we looked at the stances per keyword groups. As the large changes per month complicated the interpretation, we looked at the yearly change. Average stance per year was relatively stable across topics and their ranking changed little. Most negative keywords for both outlets were keywords often used in relation to radical-right and liberal opposition, like "multiculturalists", or "globalists" and secondly the topics related to race. The migration-related keywords that were the most prominent in our data, and therefore contributed the most to the overall changes in stances, ranked in the middle for both of the news sources. Findings suggest that from the types of groups in our filtered dataset, the more negative framing is firstly for race, then nationalities and only thirdly migrants more generally. This finding should be approached with caution as it is based on an average over years and with relatively fixed keywords. Furthermore, the stances per keyword topic are often more nuanced than our current approach can distinguish (Koppel & Jakobson, 2023).

**Stance per keyword group events**

In relation to media events, 2018-2019, for Uued Uudised, all of the keyword groups, except the xenophobia and multiculturalism related one, took a more negative stance. For Ekspress, there were more differences per topic. Smaller increase in general anti-immigration stance came mostly from the keywords related to the large topic of migration in general. There were also topical differences relating to the Ukrainian war in 2022. In Uued Uudised the relative amount of negative sentences about immigration increased but many other topics got less negative, potentially showing a shift in focus towards the war and refugees. In Ekspress Group, there is a noticeable sharp decrease in Against stances related to radical-right liberal opposition.

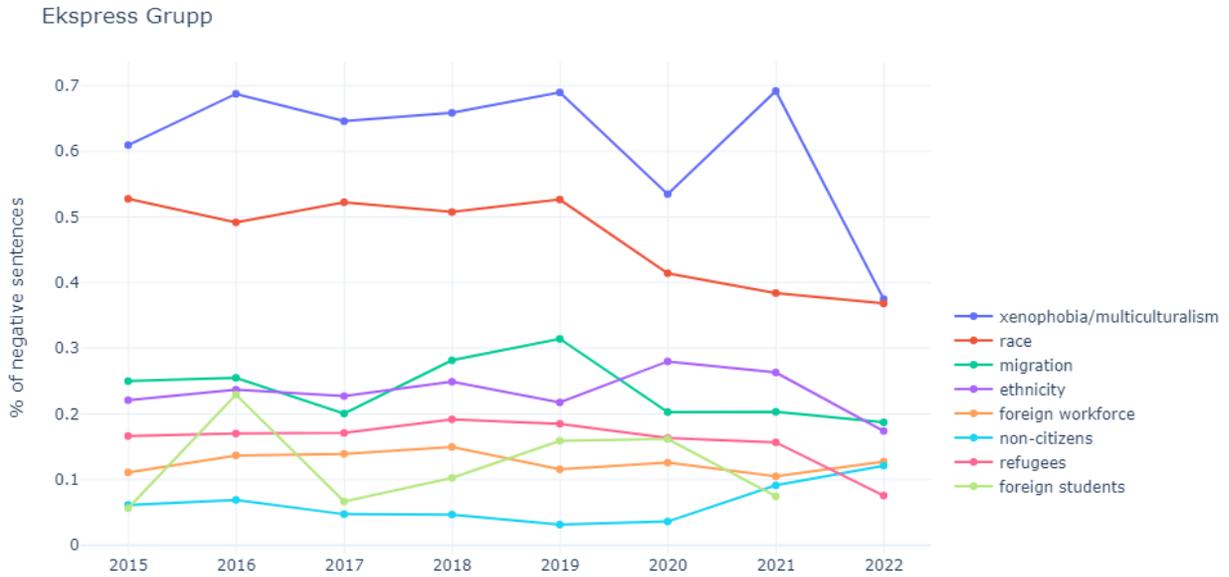

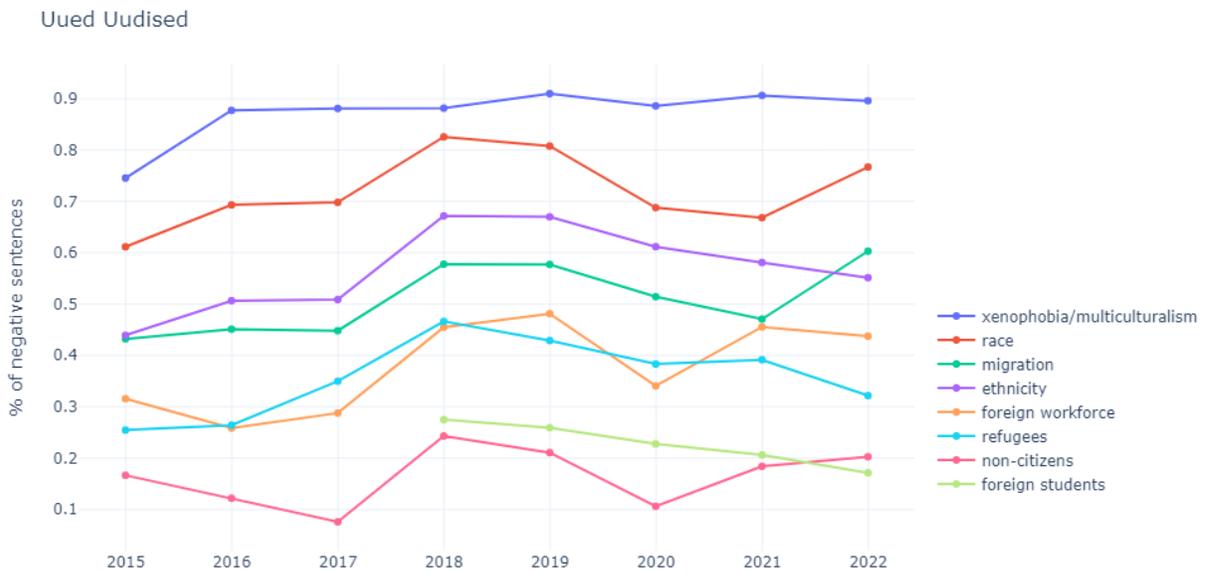

**Figures S6 and S7. Changes in Against stance per keyword group.** Shows the yearly changes per keyword group. We found the Against stance most informative for analyzing the changes dependent on specific keyword groups. Notice the different scale of y axis.

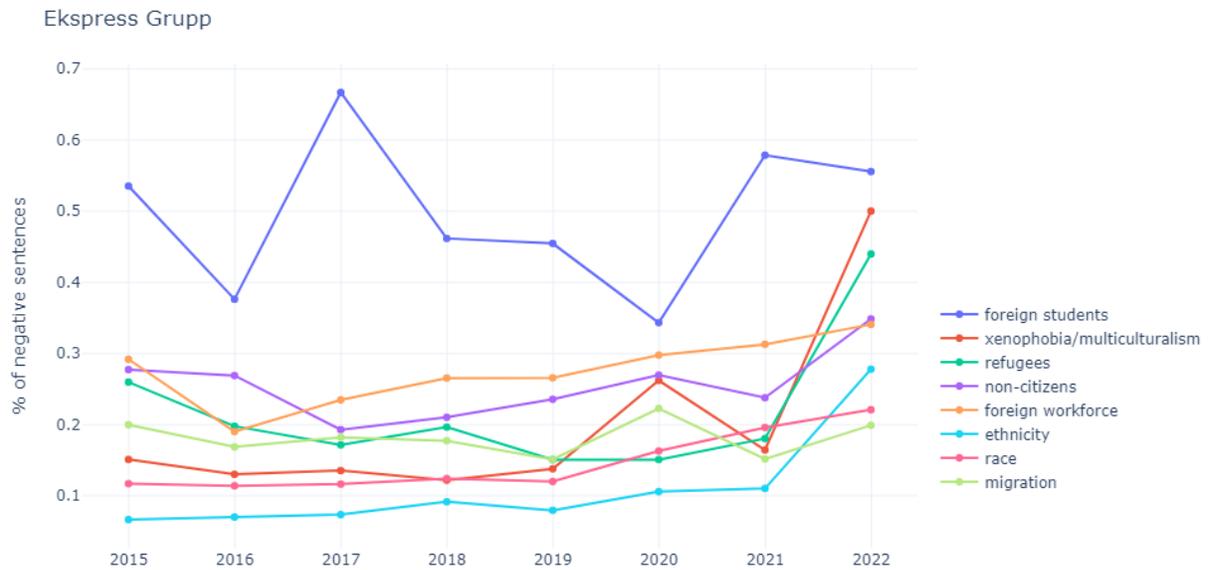

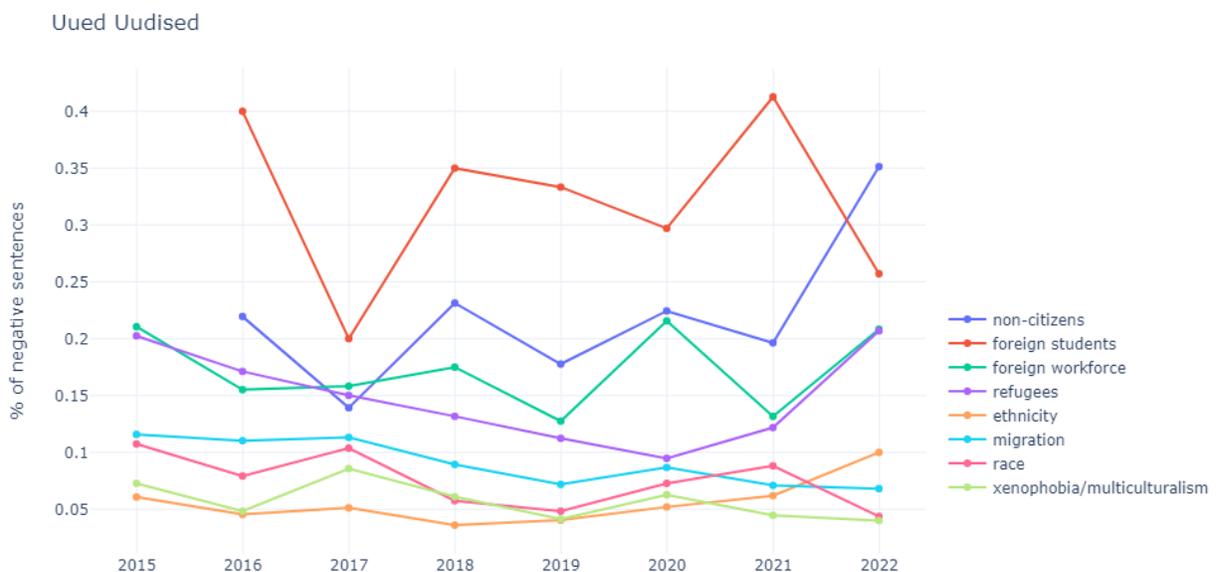

**Figures S8 and S9. Changes in Against stance per keyword group**. Depicts yearly changes. For Ekspress Grupp in 2022 all topics, except foreign students, are getting more positive. Especially xenophobia\multiculturalism related keywords and the large refugee keyword group. Interestingly, there is a slow increase of supportive stances towards foreign workforce across time. There are much smaller changes in 2022 for Uued Uudised for Uued Uudised. But there is more differentiation between topics.

## 11. Stance trends with threshold

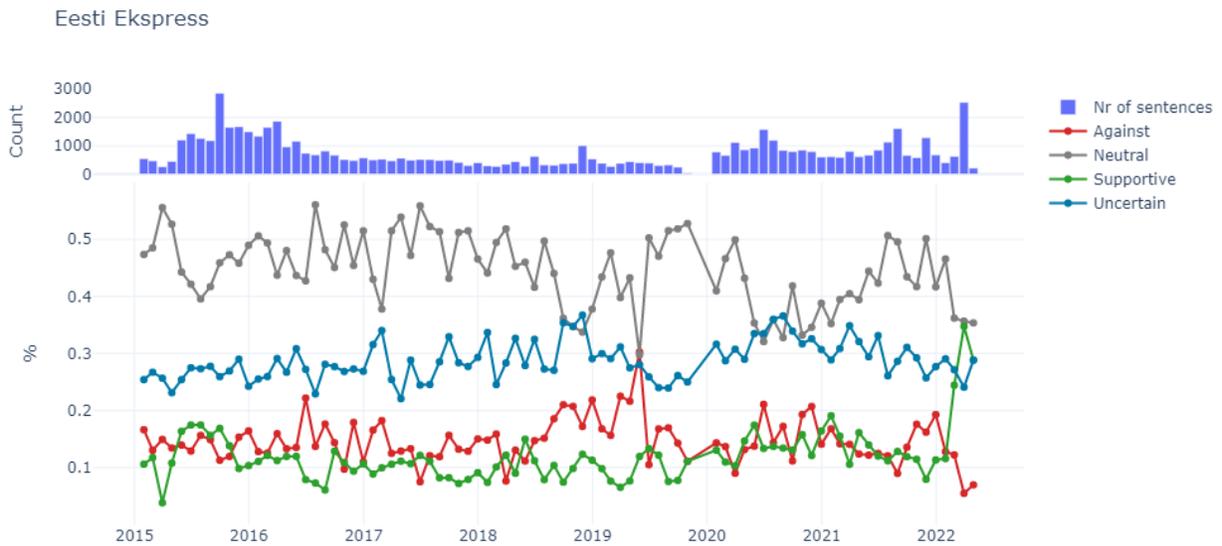

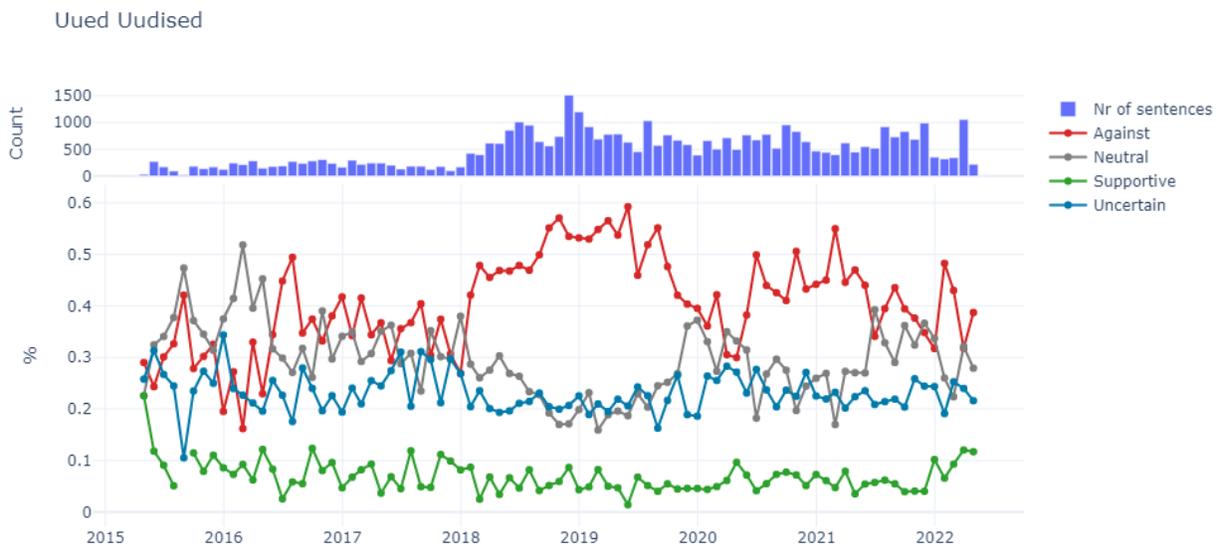

**Figures S10 and S11. Relative amount of stances plus a group of stances that were less uncertain**. The uncertain class contains sentences that have less than 70% probability of fitting into a specific class. E.g. a sentence under threshold may have 65% probability of being anti-immigration and the other 35% is shared between neutral and pro-immigration class. The plot shows that using thresholds does not have a large impact on the general trends. Below threshold sentences is relatively larger in Ekspress Grupp than in Uued Uudised, but overall difference is from 5-10% and this variability is relatively small across months.

## 12. Sentence embeddings trends

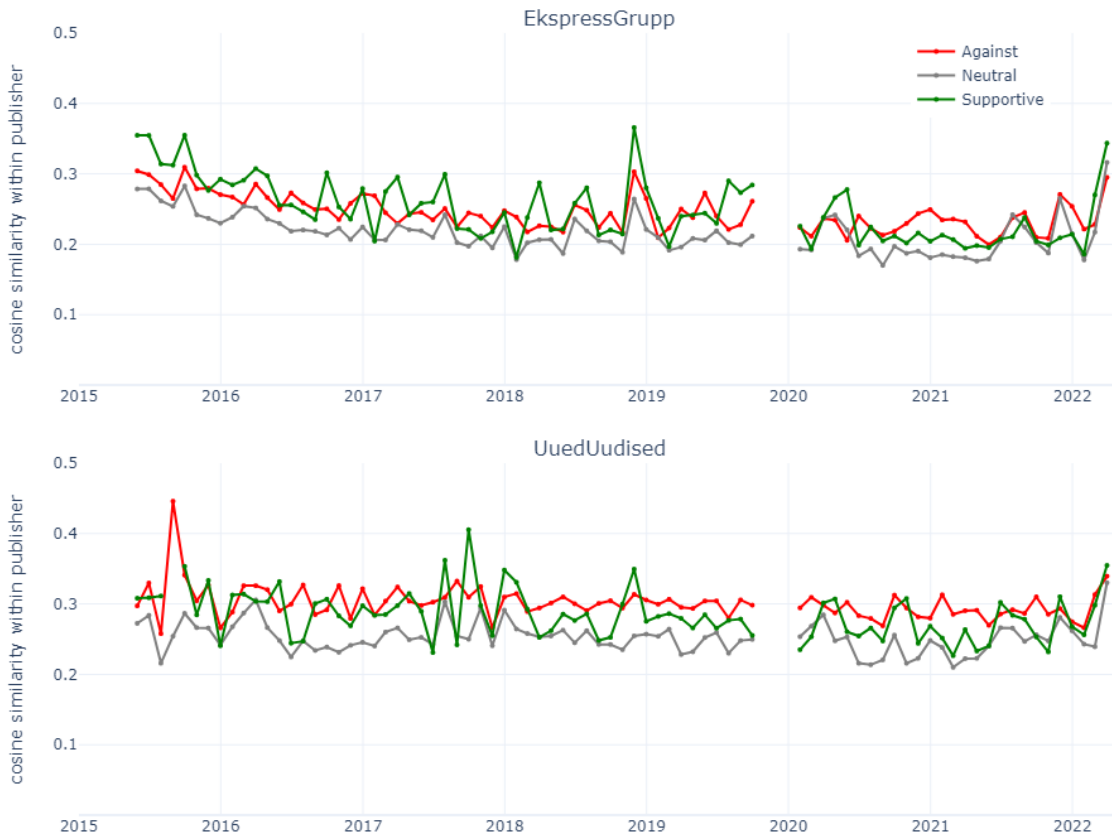

**Figure S12. Monthly average cosine similarities of sentences within the same.** Similarities are calculated separately per stance, e.g. comparing neutral sentences to other neutral sentences.